\documentclass[10pt,twocolumn,letterpaper]{article}

\usepackage{iccv}
\usepackage{times}
\usepackage{epsfig}
\usepackage{graphicx}
\usepackage{amsmath}
\usepackage{amssymb}

 \pdfoutput=1 

\usepackage[pagebackref=true,breaklinks=true,letterpaper=true,colorlinks,bookmarks=false]{hyperref}

\iccvfinalcopy 

\def\iccvPaperID{1703} 

\ificcvfinal\fi

\begin{document}
 \nocite{*}
\title{Re-designing cities with conditional adversarial networks}

\author{Mohamed R. Ibrahim$^1$, James Haworth$^1$, Nicola Christie$^2$\\
$^1$SpaceTimeLab, University College London (UCL)\\
$^2$Centre for Transport Studies, University College London (UCL)\\
London, UK\\
{\tt\small \{mohamed.ibrahim.17, j.haworth, nicola.christie\}@ucl.ac.uk}

}
\maketitle
\ificcvfinal\thispagestyle{empty}\fi

\begin{abstract}
This paper introduces a conditional generative adversarial network to redesign a street-level image of urban scenes by generating 1) an urban intervention policy, 2) an attention map that localises where intervention is needed, 3) a high-resolution street-level image (1024 X 1024 or 1536 X1536) after implementing the intervention. We also introduce a new dataset that comprises aligned street-level images of before and after urban interventions from real-life scenarios that make this research possible. The introduced method has been trained on different ranges of urban interventions applied to realistic images. The trained model shows strong performance in re-modelling cities, outperforming existing methods that apply image-to-image translation in other domains that is computed in a single GPU. This research opens the door for machine intelligence to play a role in re-thinking and re-designing the different attributes of cities based on adversarial learning, going beyond the mainstream of facial landmarks manipulation or image synthesis from semantic segmentation. 
\end{abstract}

\section{Introduction}


\begin{figure}[t]
\includegraphics[width=0.93\linewidth]
                   {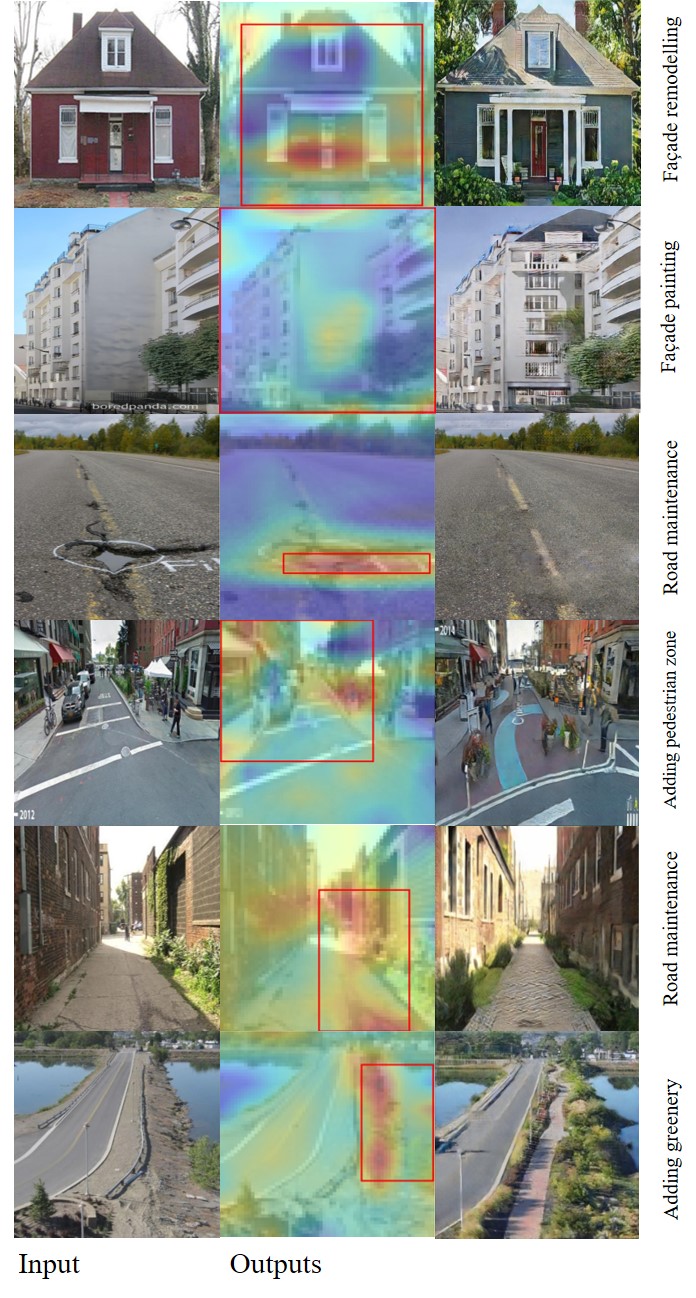}
\centering
\caption{The input and outputs of our proposed DesignerGAN method  trained as a single model for the different urban intervention tasks}
\label{fig:output}
\end{figure}
Cities change rapidly and technical surveys and inspections for maintenance,  identification of  infrastructural issues or re-modelling urban areas requires expert knowledge and high costs and resources.  Assessing the impact of urban policies before their implementation by rendering and visualising their effects in a realistic urban setting remains an expensive and challenging task for urban planners and policymakers. Usually, virtual 3D models are built and rendered to mimic the built-environment and simulate the adopted policies for intervention [1]. While this approach could lead to understanding the impact of the adopted policies, it also requires expert knowledge, time and resources to implement. Furthermore, its implementation remains limited to a given context or a project, without the ability to transfer and adapt to other contexts.  

Learning to design by machines has shown a substantial leap, most significantly, in the last five years. Generative models, specifically Generative Adversarial Networks (GANs) have shown progress in generating synthesised images of high resolution, realism, and variation [2], [3]. For instance, GANs have been utilised to generate and render facial landmarks [4], [5], semantic segmentation of urban scenes [6], [7], visualising the impact of global warming in cities [8], or even to designing architectural floor plans from zoning diagrams [9]. In this paper, we introduce a new model and dataset to generate a street-level image after applying an urban intervention, in addition to generating the policy for intervention and an attention map for localising where the intervention is needed in the input image. The model architecture comprises three components: 1) a conditioned generator, 2) a policy classifier, and 3) a discriminator. It has been trained based on a new objective loss that utilises a  $\min\max$ loss. In a single trained model in a single GPU, the introduced model is capable of changing various attributes of street-level images and applying multiple urban interventions when needed such as façade remodelling, façade painting, adding greenery, road maintenance (i.e. adding pavement, removing potholes, removing treefalls or removing garbage), road remodelling (i.e. adding pedestrian-only zones, adding road calming solutions), remodelling urban open spaces, or adding cycle lanes (See Figure 1). 

After the introduction section, the paper is structured as follows: 1) a review of relevant methods related to our work, 2) the architecture of the introduced method, DesignerGAN, and its objective loss, 3) Introducing the new dataset, and the conducted experiments, in addition to the base model and evaluation metrics used to evaluate our results, 4) presenting our results, 5) discussing our findings in the context of the current literature and the future work needed to advance this research, and last 6) a summary and conclusion of our work.


\section{Related work}

Advances in image synthesis have been achieved by GANs, in which two models: Generator (G) and Discriminator (D) are trained simultaneously in a $\min\max$ game to generate conditioned or unconditioned results. Since its first development to generate syntheses of greyscale MNIST digits [10], GANs have been developed and expanded to accommodate the synthesis of naturalistic images of high resolution and high-variance, outperforming other methods such as autoencoders, and restricted Boltzmann machines. To the best of our knowledge, there is no similar method that is used to generate images based on a condition of an urban intervention while localising where attention is needed. However, there are different pieces in the general literature of image synthesis that links with the stated issue and could offer a baseline and indicators for training and validating our method. 

\textbf{Image-to-image translation:} Most recently, machine intelligence has shown a substantial leap in image-to-image translations [7], [11], [12] and transferring a visual style from one domain to another [4], [5], [13], [14] by relying on conditional GANs (cGANs). A wide range of models has been developed to generate a conditioned realistic image synthesis. Two approaches, in particular, link with the purpose of our research. These approaches are Pix2Pix Basic [7] and HD [15], and CycleGAN [12]. The first approach, Pix2Pix, utilises cGAN through the input of paired images, where the generators generate conditioned images that are evaluated by the discriminators. The latter, CycleGAN, generates conditioned images without the need for inputting paired images. While these approaches provide best practice examples for training conditional GANs, three limitations needed to be overcome to conduct this research. First, the current methods learn only a single task for each trained model (i.e. translating a semantic segmented image to street level image or vice versa) or an image style (i.e. translating daytime image to night-time). However, unlike these existing methods, the challenges of the stated issue are threefold: 1) The complexity and the high variance of street-level images as input, especially when image translation occurs in small detail in a given image, 2) the variations of the interventions that need to be inferred and applied by the model without supervision, and 3) The applied interventions are only applied to a subset of a given image when needed, and this requires attention to detail. 

\textbf{Generative models for designing cities:} Few methods have been introduced that utilise deep generative models in remodelling cities, or the domain of urban planning and architecture in general.  The City-GAN model [16] is conducted to generate conditioned image synthesis based on a given city-style relying on cGANs networks. However, the quality of the generated images are often noisy and of low resolution (64 x 64). Furthermore, the condition is based on the overall architectural style of the image without any control over the content of the generated image. Relying on Pix2Pix HD architecture, [9] applied a cGAN model to generate an architectural floor-plan drawing from a semantic zoning diagram.  Relying on unconditioned GANs, the FaceLift model [17] is conducted to beautify a given street-level image. While the model shows potential in applyication to cities, the results are of low resolution, and the model tends to learn to alternate the overall style by manipulating the image colours rather than applying a geometric transformation to the image (for instance, the model does not remove garbage from street, but rather alternate the overall colour of the street).  Last, similar to FaceLift, [18] utilised GANs model to beautify cities. While the results are significantly better than the ones introduced in FaceLift [17], the model tends to match a given input image to the closet one in the dataset. As a result, the introduced method is rather a matching tool of two different images of a similar pose than a model that generates an image by changing a given urban attribute in the original street-level image. 

To achieve the objectives of this research, our method requires performing multiple tasks (such as remodelling façade, road, adding tress, removing potholes, etc.) in a single model. Second, Geometric transformation is key to the success of the introduced method whereas, for instance, CycleGAN performs well with style transfer but suffers with complex geometric transformation [12]. Thus, these issues add additional challenges for training and inference. Last, the current methods are evaluated based on their perceptual appearance. While the progress in realistic output is seminal for the advancement of GANs, our introduced model not only seeks a realistic output but also evaluates how a given task is learned. For instance, a noisy background, in the case of our model, should be treated with less attention to whether the model learns to remove a pothole when needed.

\section{Method}

\subsection{DesignerGAN architecture}

Figure 2 shows the overall architecture of the introduced method. It comprises three main components: Generator, discriminator, and policy classifier. 

\textbf{The Generator ($G$)} is based on the U-Net architecture [19]. However, three major changes have been introduced. First, it comprises an encoder and decoder of three down-sampling and up-sampling convolutional blocks, respectively. Each convolutional block comprises a convolution layer, followed by spatially-adaptative normalisation, as introduced in [20] and max-pooling layers, activated by a Leaky ReLU function. Second, the number of feature maps of the four convolutions of the encoder part are 128, 32, and 16 respectively, whereas the number of feature maps of the four convolutions of the decoder reciprocate the ones of the encoder. Based on experiments, we have doubled the kernel size of the outer layers, which has proven to be a more effective strategy than adding extra layers. Second, similar to [7], we also added residual skip connections [21] between the matching convolution layers in the encoder and decoder to concatenate their channels (layer j and layer n-j, where n is the total number of layers).

\textbf{The Discriminator ($D$)} is based on three residual blocks of the down-sampling convolutional structure. Similar to the encoder part of G, each block comprises a convolution layer followed by batch normalisation and a max-pooling layer activated with a ReLU function. However, each convolutional block comprises a skip connection  [21] by adding its input to the output of the block.  The number of features in the three blocks are 128, 32, and 16 respectively with a kernel size (3 X 3), (3 X 3) and (1 x 1) strides and zero paddings. 

\textbf{The Policy classifier ($Q$)} represents a second branch. We used four convolution blocks, in which each block comprises two convolutional layers, followed by a batch-normalisation and max-pooling layer of kernel size (2 X 2). The number of feature maps of the convolutional layers is the same for the two convolutional layers in each block, and double the value of the previous block, as follows 16, 32, 64, and 28 respectively, whereas their kernel size is (3 X 3) for all convolutional layers. The four blocks are followed by a global average-pool layer and a fully-connected dense layer of N neurons, where N is the number of intervention policies. All layers are activated by a ReLU function, except for the last one which is activated by a Softmax function. In addition to classifying a given policy, we also generated a coarse localisation map that highlights the attention which influences the predicted class by utilising Gradient-weighted Class Activation Mapping (Grad-CAM) [22].

\begin{figure}
\centering
\includegraphics[width=0.85\linewidth]{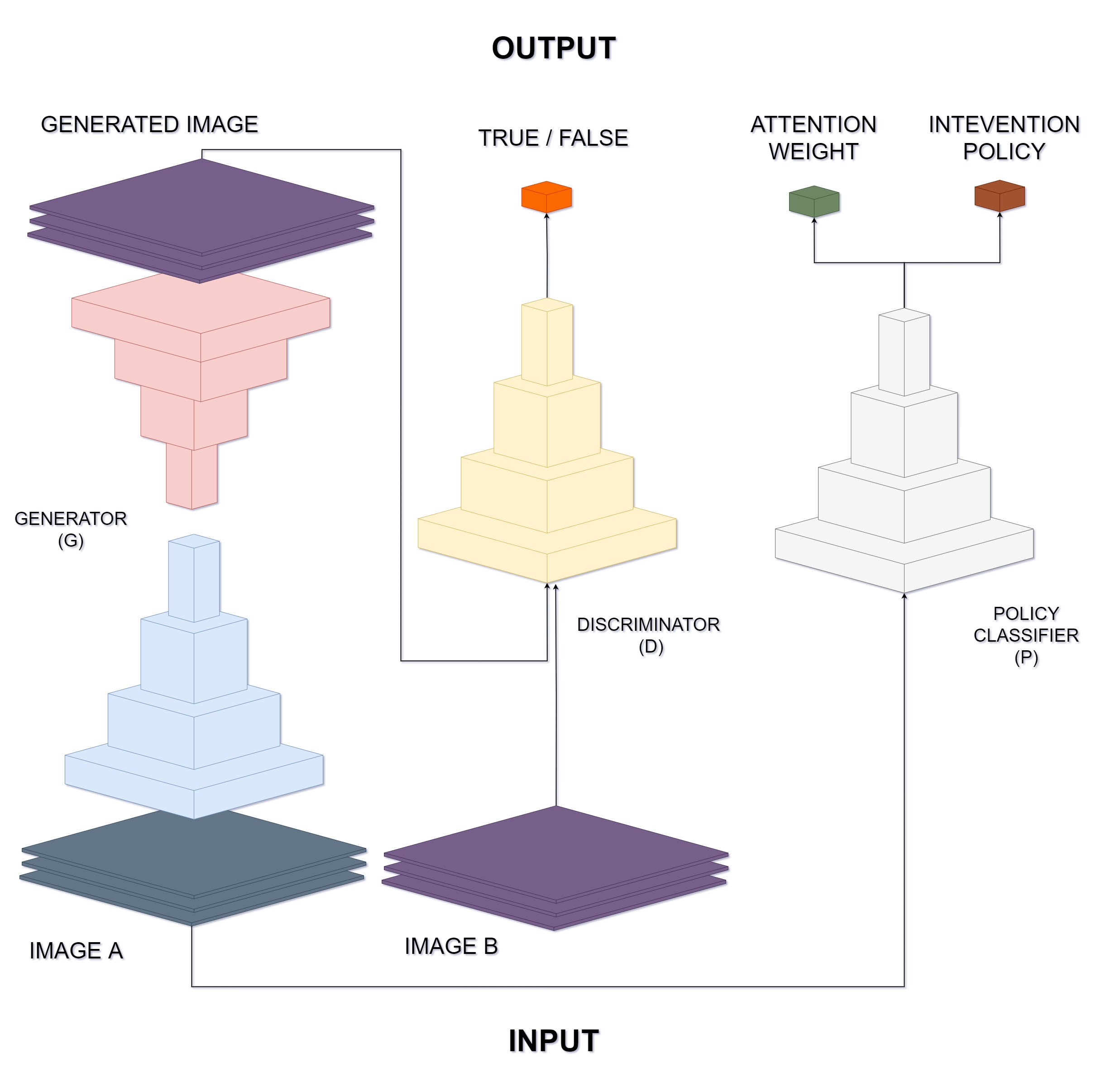}
\centering
\caption{The architecture of DesignerGAN}
\label{fig:architecture}
\end{figure}

\subsection{Objective loss}

The overall objective of the introduced model is based on a $\min\max$ loss where $G$ attempts to minimise the policy and the generator losses against $D$ that attempts to maximise it, in addition to generating images that are as close as possible to the ground truth. Additionally, we have added further penalisations, in which $G$ is not only penalised on the overall generated image, but also in the absolute difference between the input and output, to give additional weight to the region of interest of the image where intervention is needed. We also added $L_1$ regulation as it is proven to enhance the quality of the generated image[7], [23]. The overall loss is defined as: 

\begin{equation} \label{eq1}
\begin{split}
Loss &=\arg{{\min}_G\ {\max}_D}L_{cGAN}\left(G,D\right) \\
     & + \arg{{\min}_P\ L_{policy}\left(Q\right)+\ {w}{L_1}(G)}
\end{split}
\end{equation}

Where:
\begin{multline*}\label{eq2}
L_{cGAN}(G,D) =\mathbf{E}_{x,y}[\log D{(x, y)}]+\mathbf{E}_{x,y}[\log D{(||x -y||)}]\\
+\mathbf{E}_{x,z}[\log(1- D(x,G(x,z))]
\end{multline*}

\[L_{policy}(Q_{x,m}) = - \sum_{i=1}^{N}x_{i} \log (m_{i}) \]

\[L_{L1} (G) =\mathbf{E}_(x,y,z)\ [||y-G(x,\ z)||]\]
given that the model input is based on paired images and an intervention policy $(x, y, m)$, in addition to a Gaussian noise $z$ [24], N is the total number of intervention policies, and $w_1$ is a given weight.

\section{Experiments}

\subsection{Dataset}
To the best of our knowledge, there are no datasets that combine urban interventions for street-level images for a given urban area before and after the implementation of a given policy. Accordingly, building our dataset is the only way to conduct this research. 

\textbf{Re-design dataset:} We introduce a new dataset that comprises paired street-level images of architectural and urban design remodelling projects. This dataset includes 372 paired images (744 in total) whereas images are aligned based on before and after applying a given urban policy. The images are collected from the web from different cities across the globe based on real-life urban interventions. The paired images have an identical camera pose, and all images are taken in daytime and clear weather. After collecting the images, we have inspected their quality and alignments. All images are resized to  1536 X 1536 with RGB colour channels. Furthermore, we sorted them into eight policy categories such as 1) adding cycle lane, 2) adding greenery, 3) adding pedestrian-only zone, 4) adding sidewalks,  5) façade painting, 6) façade remodelling, 7) open space remodelling (i.e. adding pedestrian pathway, re-modelling green spaces), and 8) Road maintenance (i.e. garbage removal, road pavement, pothole removal). We used this data for training, testing and validating our method. Table 1 shows the count of images per policy category. Figure 3 shows a sample of paired images for each policy category. See supplementary materials section, for further information regarding the definition of each policy class, collection methods, and how to access the dataset online.

\begin{table}[h]
\begin{center}
\begin{tabular}{|l|c c |}
\hline
ID & Policy $(m)$& Number of images\\
\hline\hline
1& Adding cycle lane &24 \\
2 & Adding greenery &40\\
3&	Adding pedestrian only zone&54\\

4&	Adding sidewalks&55\\
5&	Façade painting&23\\
6&	Façade remodelling&97\\
7&	Open space remodelling&29\\
8&	Road maintenance&50\\

\hline
\end{tabular}
\end{center}
\caption{Dataset classification based on policy categories  }
\label{tab:hresult}
\end{table}

\begin{figure}[t]
\includegraphics[width=1\linewidth]
                   {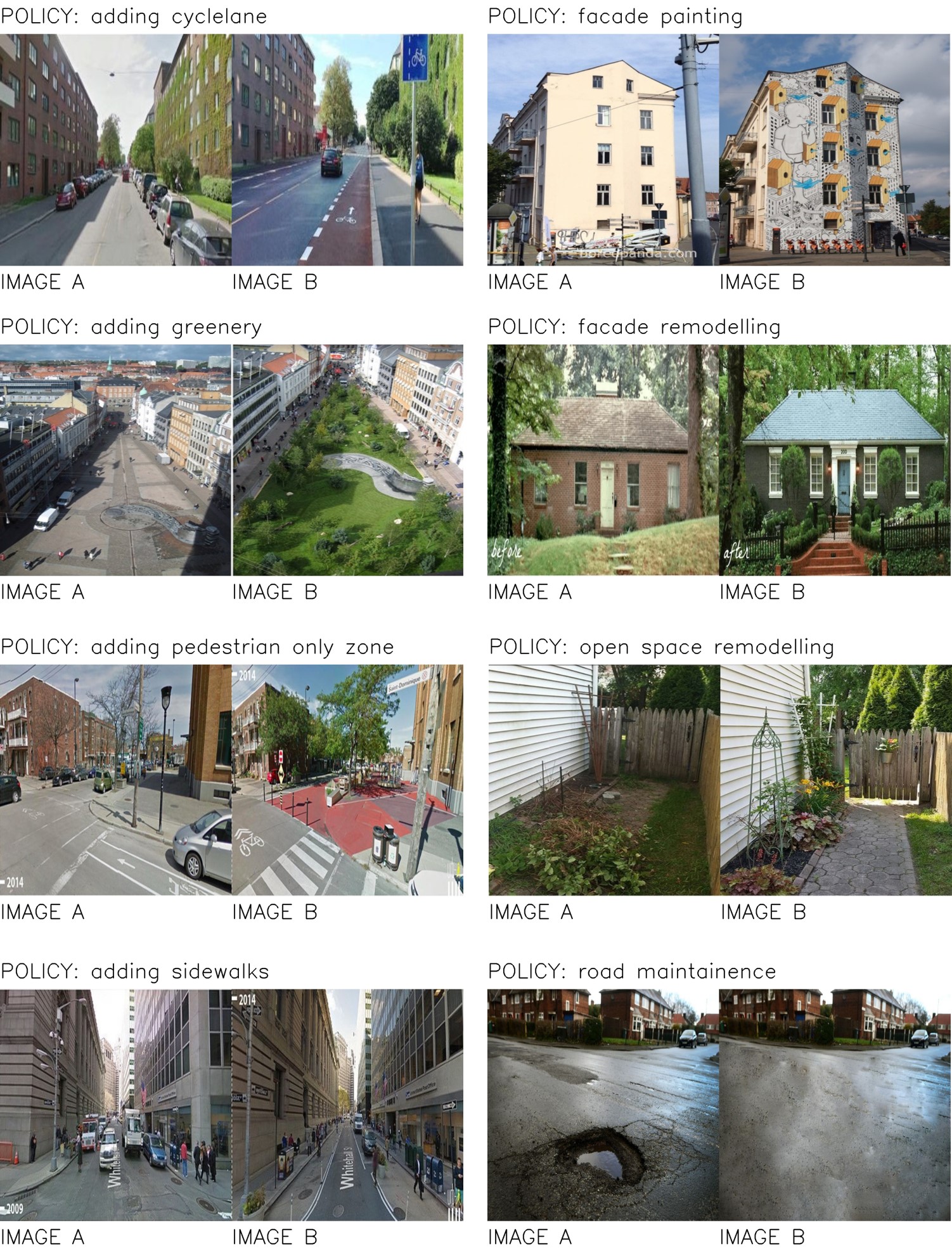}
\centering
\caption{A sample of the dataset}
\label{fig:sample}
\end{figure}

\subsection{Base model and evaluation metrics}

It remains a challenge to evaluate the introduced method against any existing methods due to the absence of benchmark datasets and methods used for the same purpose. However, the closet existing method for evaluating our generated image is Pix2Pix [7], which utilises cGAN and is trained on paired images, nevertheless, the Pix2Pix model is optimised using only a single GPU. Accordingly, we have trained a Pix2Pix model in our dataset to give a baseline indicator of how our introduced method performances against the existing ones. 

Evaluating the realism and quality of the generated synthesis remains on-going research [2], [3]. However, we followed the common protocols by using the Frechet Inception Distance (FID) score [25] that is computed based on the pooling layer of an  Inception v3 network [26]. It is defined as: 

\begin{equation} \label{eq5}
FID=||\mu_r-\mu_g||^2 + T_r (\sigma_r+ \sigma_g -2(\sqrt{\sigma_r \sigma_g }))
\end{equation}
given that $\mu_r$ and $\mu_g$ are the mean of the features of the real and generated images, $\sigma_r$ and $\sigma_g$ are the covariance matrices and  $T_r$ represents the trace operator relying on Inception v3 network.


Furthermore, we evaluate our method based on two new indicators that serve the scope of the introduced method. First, we used the trained policy classifier to generate a classification probability for both the generated image and its ground truth to judge the quality of the generated image, assuming that the closer the generated image to the ground truth, the smaller the difference between the predicted probabilities of both generated and ground truth images. Accordingly, it becomes a regression problem, which can be evaluated relying on MSE loss. Second, while a realistic perception of the generated image is important, evaluating the applied intervention remains a crucial issue- for the scope of this research. Accordingly, we used the FID score to evaluate the region of interest by finding the absolute difference between the input and ground truth images (image A and image B) to the difference of the generated image, and its input (image A, generated B). By doing so, the model is penalised and evaluated based on the region of interest that shows the transformation, instead of the entire image (See Figure 4). 

\begin{figure}[t]
\includegraphics[width=0.8\linewidth]
                   {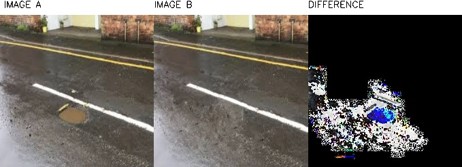}
\centering
\caption{Region of interest: $||x-y||$}
\label{fig:ROI}
\end{figure}


\subsection{Training and optimisation }
The method is trained based on stochastic gradient descent relying on Adam optimiser [27] with a batch size of 1, an initial learning rate of 0.0002, and momentum parameters of 0.5 and 0.9 for $\beta_1$ and $\beta_2$ respectively, whereas a linear learning rate decay is applied after the first 200 epochs to assist in optimisation and generalisation. The three parts of the model ($G, D, P$) are trained based ensemble learning [28] where ($G,D$) are trained for 600 epochs simultaneously to optimise a step of the gradient descent of $D$ and $G$ respectively, and followed by training $P$ for 25 epochs. Note that the three components of the model can be trained also simultaneously as a multi-task model, but that would require multiple GPUs of a larger shared memory. At inference, the $G$ and $P$ components are assembled and the trained weights output a predicted urban intervention policy, attention map, and a generated realistic street-level image with the implementation of the predicted policy.

\section{Results}
We trained two models of image input size of (1024 X 1024 X 3) and (1536 X1536 X 3) respectively. In Figure 5, we show the results of the first model. It shows the three outputs of the introduced method, which are the generated attention map, predicted intervention policy, and image synthesis after an intervention. The figure shows unequivocally how the introduced method learned to localise attention to the region where intervention is needed without supervision in localisation. Also, the figure shows a high degree of realism of the generated image after intervention in comparison to the ground truth images. 

Figure 6 shows a wide range of image syntheses generated by our introduced model (1024 X 1024) and the Pix2Pix. By visual inspections, the results of our model surpass the generated images by the Pix2Pix model in all types of intervention policies. The figure shows that the images generated by Pix2Pix are suffering from a limitation of fine details, in addition to over-fitness in the learned features that lead to pattern repetition across different input images regardless of the type of intervention needed.

Besides inspecting the generated images based on their visual appearance, we also computed the FID score to quantify the quality of the generated image to their respective ground truth. Table 2 shows the results of FID scores for both the introduced models and the selected baseline model. It shows how our models outperform the Pix2Pix model. Nevertheless, it shows that the model of images of a larger input size (1536 X 1536) surpasses the one with an input size of (1024 X 1024), indicating that more small details can be learned, which increases the overall realism of a generated image. Also, the table shows the Crossentropy loss of the policy classifier during testing, highlighting a high validation of learning a given intervention policy. 

While training a model with a larger input size could lead to better results, it comes at expense of the computation required to optimise and train the model. Table 3 shows the number of hours needed to train both models in a single Titan V GPU. A slight improvement in the generated images (in term of the FID score) comes at the expense of a substantial increase in the computation time for training (approximately 50 percent increase of the original time).

\begin{table}
\begin{center}
\begin{tabular}{|l|c c|}
\hline
Method & FID score $\downarrow$ & CE  $\downarrow$\\
\hline\hline
Pix2Pix & 234 & -\\
Ours (1024 X 1024) & 66 & 0.011 \\
Ours (1536 X 1536)  &68 & 0.024\\
\hline
\end{tabular}
\end{center}
\caption{Evaluation metrics based on FID score for generated images and categorical Crossentropy (CE) for generated policy class}
\end{table}

\begin{table}
\begin{center}
\begin{tabular}{|l|c c|}
\hline
Model input & GPU &Training time  \\
\hline\hline

1024 X 1024 & 1 (Titan v) & 41 hours (1.7 days)\\
1536 X 1536 &1 (Titan v) &86 hours (3.5 days)\\
\hline
\end{tabular}
\end{center}
\caption{Training time for the different methods}
\end{table}

\begin{figure*}
\begin{center}
\includegraphics[width=0.7\linewidth]{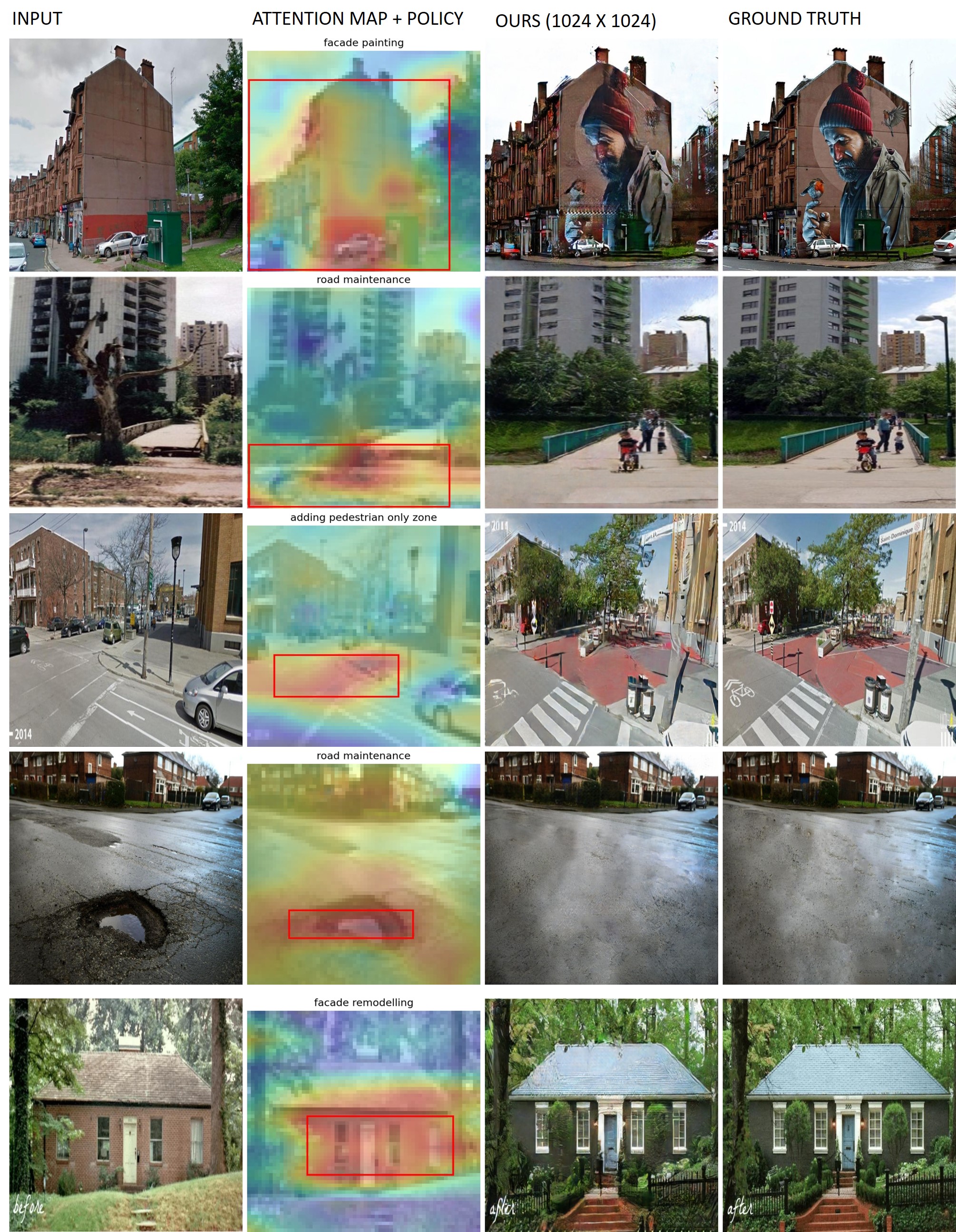}
\end{center}
   \caption{Proposed method results.}
\label{fig:results1}
\end{figure*}


\begin{figure*}
\begin{center}
\includegraphics[width=0.7\linewidth]{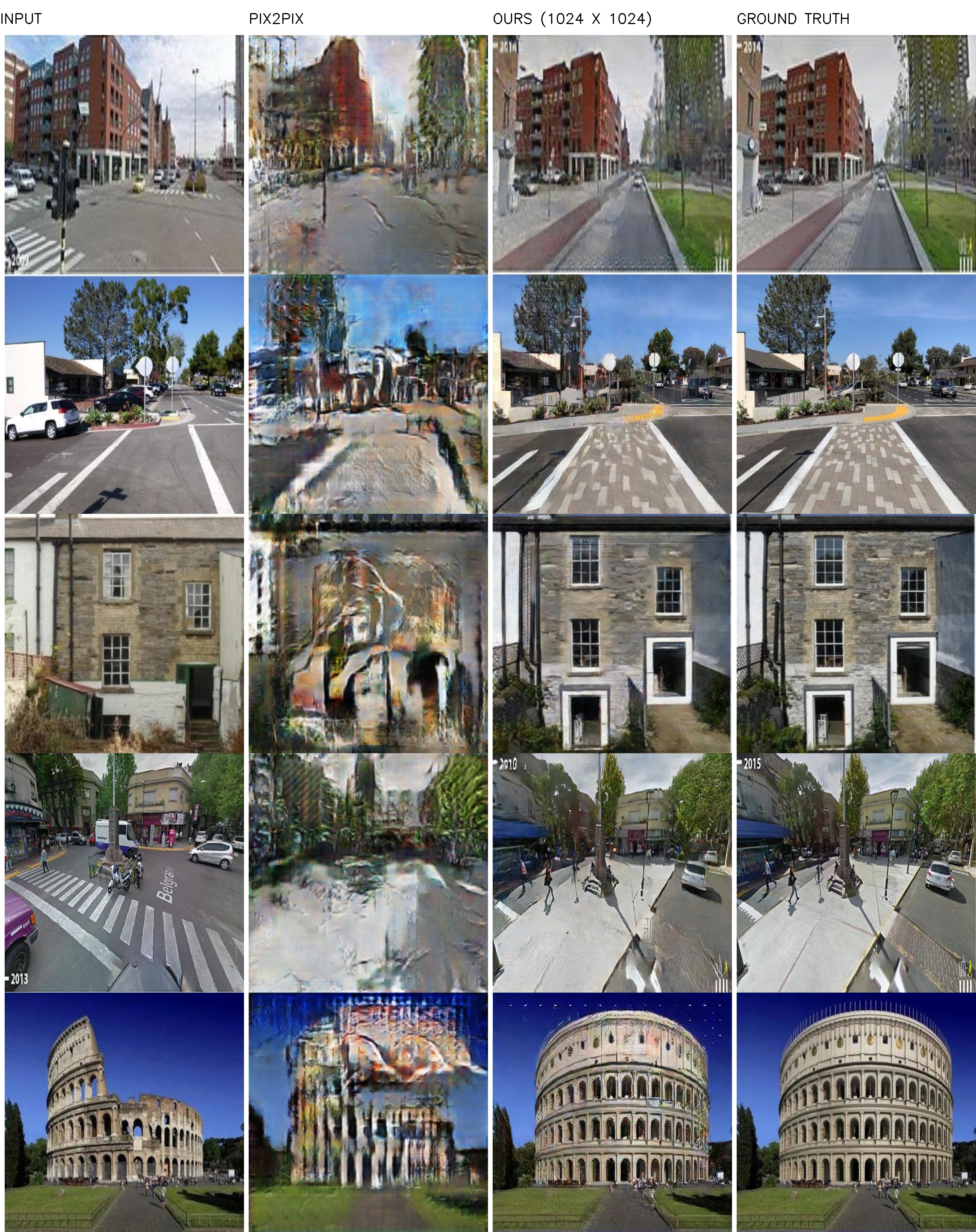}
\end{center}
   \caption{Comparing our results (1024 X 1024) with Pix2Pix model.}
\label{fig:results2}
\end{figure*}


\section{Discussion}
\subsection{Re-designing cities with AI-generated policies}

Cities are continuously and rapidly changing and generating intervention policies for maintenance and re-configuration for a given urban area remains expensive and requires expert knowledge. Visualising policies to show their impact on urban areas with a high level of realism remains a challenge. In this paper, we introduced a new dataset and method that simultaneously generates an intervention policy for an urban area at street-level, while localising the attention where the intervention is a need, in addition to generating a high-resolution street level-image after implementing this generated intervention. Figure 7 shows an input-output example of our introduced method. 

In this paper, we show a new approach for intervening in urban areas and reconfiguring their attributes by intervention policies beyond the main streams of utilising GANs models in generating and altering facial attributes that have a limited number of features and complexities when compared to street-level images. We introduce a single GAN model that is trained on multiple conditions to output a realistic image from a realistic image beyond the common stream of translating images from semantic to realistic outputs or vice versa in street-level images such as cityscape dataset [6]. 
\begin{figure}[t]

\includegraphics[width=0.85\linewidth]
                   {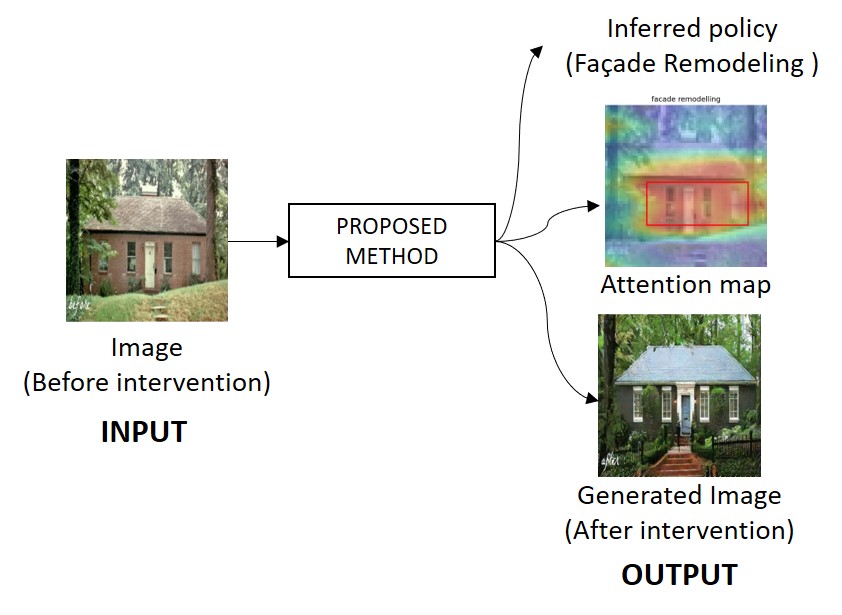}
\centering
\caption{Proposed model}
\label{fig:ai_policy}
\end{figure}

\subsection{Computational resources vs perceptual improvement}

It is evident that an increase in the amount of computational resources is associated with a perceptual improvement in training image synthesis based on adversarial learning. Nevertheless, the depth and the number of parameters of the introduced method is a trade-off between the available shared memory and the expected improvement in the quality of the image synthesis.  For instance, the  Pix2Pix HD model [15] is trained, as stated in their paper, on a GPU with 24GB of shared memory to reach a resolution of 1024 X 1024. The StyleGAN [4] and StyleGAN2 model [5] are trained on 8 Tesla V100 GPUs to output image synthesis of resolution 1024 X 1024, achieving a top FID score of  4.40 and 2.83 respectively. While these substantial efforts are seminal for improving the quality of the image synthesis and the advancement of the field in general, the challenges remain in achieving high-resolution results with minimal costs and computational resources. In this paper, we attempt to show evidence of generating images at a high resolution (up to 1536 X 1536) with limited computational resources of a single GPU. We also have shown evidence that the introduced method achieves stronger results than the counterpart baseline methods that rely on a single GPU for training image synthesis.  
\subsection{Inferring attention without attention}
There are several GANs architectures that utilise context-aware or self-attention layers for image generation [29]–[32]. We have experimented with adding self-attention layers to allow our model to understand the region of interest where intervention is needed the most. We found that adding a self-attention layer comes at the expense of shared memory, and the required time for training and inference, as it has increased the number of trained parameters of the model which could not train with limited shared memory of 12 GB on a single GPU. Alternatively, we have utilised Grad-CAM [22] as previously introduced for localising attention in image classification tasks by structuring and visualising what has been learned in the convolution structure during training. This implementation only occurs post-training at the inference, in which this approach has shown effective results in the purpose of our model for showing context-aware and localised attention with a minimal number of parameters.
\subsection{Data size, the variation of conditions, and the effect on training}
Several previous works have shown the potential of cGAN in learning a new task and generating high-quality image synthesis with a minimal number of trained images. For instance, [7] have used cGAN with a dataset of only 91 webcam images reporting a good result in generating night-day image translation based on an aligned conditioned input. This observation is also in line with our findings, however, after training two models; one with a subset of images (100 images less) and the other with the full dataset, we observed an increase of FID score with 6 percent by adding those 100 images, for the same image resolution, policy classes, and training implementations and hyperparameters. 

On the other hand, most of the existing methods that utilise cGAN to generated a conditioned output are only based on learning a single condition, usually, a semantic representation [11], [12], [15], [20], [33]–[35], whether or not a latent space that generates a variation for a given input image is included. Nevertheless, several models of the same method are trained on different datasets on different tasks to give an indication of the versatility and generality of the introduced method. While the variance in images in these methods remains high, the variance in the learned condition is low. In contrast, our introduced method is trained on a dataset of multiple tasks (intervention policies) that increase complexity in the learned variance of images and the variance in the learned tasks. Nevertheless, the translation between a realistic image to a realistic image occurs at a subset of the dataset that requires higher attention. We have observed that the increase of several conditions represented in the number of intervention policies comes at the expense of the complexity of the learned parameters, training stability, and accordingly in the expense of the quality of the generated images. However, it increases the versatility of the introduced method in learning multiple conditions in a single trained model, unlike the common stream of the introduced methods that train several models on several datasets of a single condition. 
\section{Conclusion}
Generating and visualising urban policies and interventions in realistic scenes remains an expensive task. It requires experts’ knowledge, and resources for planning, modelling, and rendering. In this paper, we introduced a new model to re-design cities by 1) generating a policy for intervention, 2) localising areas that need attention and 3) generating a synthesis of street-level images showing the implementation of a given intervention at a high-resolution of up to 1536 X 1536. The introduced model utilised conditional generative adversarial learning in generating image synthesis. The model showed a high validation in learning and applying multiple policies that require high attention and geometric transformation in a single model.

\section*{References}
\setlength{\parindent}{0em}

[1]	A. Kunze, R. Burkhard, S. Gebhardt, and B. Tuncer, “Visualization and Decision Support Tools in Urban Planning,” in Digital Urban Modeling and Simulation, vol. 242, S. M. Arisona, G. Aschwanden, J. Halatsch, and P. Wonka, Eds. Berlin, Heidelberg: Springer Berlin Heidelberg, 2012, pp. 279–298.

[2]	T. Karras, T. Aila, S. Laine, and J. Lehtinen, “Progressive Growing of GANs for Improved Quality, Stability, and Variation,” ArXiv171010196 Cs Stat, Feb. 2018, Accessed: Mar. 13, 2021. [Online]. Available: http://arxiv.org/abs/1710.10196.

[3]	Z. Wang, Q. She, and T. E. Ward, “Generative Adversarial Networks in Computer Vision: A Survey and Taxonomy,” ArXiv190601529 Cs, Dec. 2020, Accessed: Mar. 13, 2021. [Online]. Available: http://arxiv.org/abs/1906.01529.

[4]	T. Karras, S. Laine, and T. Aila, “A Style-Based Generator Architecture for Generative Adversarial Networks,” ArXiv181204948 Cs Stat, Mar. 2019, Accessed: Mar. 13, 2021. [Online]. Available: http://arxiv.org/abs/1812.04948.

[5]	T. Karras, S. Laine, M. Aittala, J. Hellsten, J. Lehtinen, and T. Aila, “Analyzing and Improving the Image Quality of StyleGAN,” ArXiv191204958 Cs Eess Stat, Mar. 2020, Accessed: Mar. 13, 2021. [Online]. Available: http://arxiv.org/abs/1912.04958.

[6]	M. Cordts et al., “The cityscapes dataset for semantic urban scene understanding,” in Proceedings of the IEEE conference on computer vision and pattern recognition, 2016, pp. 3213–3223.

[7]	P. Isola, J.-Y. Zhu, T. Zhou, and A. A. Efros, “Image-to-Image Translation with Conditional Adversarial Networks,” ArXiv161107004 Cs, Nov. 2016, Accessed: Oct. 24, 2018. [Online]. Available: http://arxiv.org/abs/1611.07004.

[8]	V. Schmidt et al., “Visualizing the Consequences of Climate Change Using Cycle-Consistent Adversarial Networks,” ArXiv190503709 Cs, May 2019, Accessed: Mar. 13, 2021. [Online]. Available: http://arxiv.org/abs/1905.03709.

[9]	W. Huang and H. Zheng, “Architectural Drawings Recognition and Generation through Machine Learning,” ACADIA, p. 11, 2018.

[10]	I. J. Goodfellow et al., “Generative Adversarial Networks,” Jun. 2014, Accessed: Mar. 13, 2021. [Online]. Available: http://arxiv.org/abs/1406.2661.

[11]	M.-Y. Liu, T. Breuel, and J. Kautz, “Unsupervised Image-to-Image Translation Networks,” ArXiv170300848 Cs, Jul. 2018, Accessed: Mar. 13, 2021. [Online]. Available: http://arxiv.org/abs/1703.00848.

[12]	J.-Y. Zhu, T. Park, P. Isola, and A. A. Efros, “Unpaired Image-to-Image Translation using Cycle-Consistent Adversarial Networks,” ArXiv170310593 Cs, Aug. 2020, Accessed: Mar. 13, 2021. [Online]. Available: http://arxiv.org/abs/1703.10593.

[13]	A. Dosovitskiy and T. Brox, “Generating Images with Perceptual Similarity Metrics based on Deep Networks,” ArXiv160202644 Cs, Feb. 2016, Accessed: Mar. 13, 2021. [Online]. Available: http://arxiv.org/abs/1602.02644.

[14]	X. Huang and S. Belongie, “Arbitrary Style Transfer in Real-time with Adaptive Instance Normalization,” ArXiv170306868 Cs, Jul. 2017, Accessed: Mar. 13, 2021. [Online]. Available: http://arxiv.org/abs/1703.06868.

[15]	T.-C. Wang, M.-Y. Liu, J.-Y. Zhu, A. Tao, J. Kautz, and B. Catanzaro, “High-Resolution Image Synthesis and Semantic Manipulation with Conditional GANs,” ArXiv171111585 Cs, Aug. 2018, Accessed: Mar. 13, 2021. [Online]. Available: http://arxiv.org/abs/1711.11585.

[16]	M. Bachl and D. C. Ferreira, “City-GAN: Learning architectural styles using a custom Conditional GAN architecture,” ArXiv190705280 Cs Eess Stat, May 2020, Accessed: Mar. 13, 2021. [Online]. Available: http://arxiv.org/abs/1907.05280.

[17]	S. Joglekar, D. Quercia, M. Redi, L. M. Aiello, T. Kauer, and N. Sastry, “FaceLift: a transparent deep learning framework to beautify urban scenes,” R. Soc. Open Sci., vol. 7, no. 1, p. 190987, Jan. 2020, doi: 10.1098/rsos.190987.

[18]	T. Kauer, S. Joglekar, M. Redi, L. M. Aiello, and D. Quercia, “Mapping and Visualizing Deep-Learning Urban Beautification,” IEEE Comput. Graph. Appl., vol. 38, no. 5, Art. no. 5, Sep. 2018, doi: 10.1109/MCG.2018.053491732.

[19]	O. Ronneberger, P. Fischer, and T. Brox, “U-Net: Convolutional Networks for Biomedical Image Segmentation,” ArXiv150504597 Cs, May 2015, Accessed: Mar. 13, 2021. [Online]. Available: http://arxiv.org/abs/1505.04597.

[20]	T. Park, M.-Y. Liu, T.-C. Wang, and J.-Y. Zhu, “Semantic Image Synthesis with Spatially-Adaptive Normalization,” ArXiv190307291 Cs, Nov. 2019, Accessed: Mar. 13, 2021. [Online]. Available: http://arxiv.org/abs/1903.07291.

[21]	K. He, X. Zhang, S. Ren, and J. Sun, “Deep Residual Learning for Image Recognition,” in 2016 IEEE Conference on Computer Vision and Pattern Recognition (CVPR), Las Vegas, NV, USA, Jun. 2016, pp. 770–778, doi: 10.1109/CVPR.2016.90.

[22]	R. R. Selvaraju, M. Cogswell, A. Das, R. Vedantam, D. Parikh, and D. Batra, “Grad-CAM: Visual Explanations From Deep Networks via Gradient-Based Localization,” in International Conference on Computer Vision (ICCV), 2017, p. 9.

[23]	D. Pathak, P. Krahenbuhl, J. Donahue, T. Darrell, and A. A. Efros, “Context Encoders: Feature Learning by Inpainting,” ArXiv160407379 Cs, Nov. 2016, Accessed: Mar. 14, 2021. [Online]. Available: http://arxiv.org/abs/1604.07379.

[24]	X. Wang and A. Gupta, “Generative Image Modeling using Style and Structure Adversarial Networks,” ArXiv160305631 Cs, Jul. 2016, Accessed: Mar. 14, 2021. [Online]. Available: http://arxiv.org/abs/1603.05631.

[25]	M. Heusel, H. Ramsauer, T. Unterthiner, B. Nessler, and S. Hochreiter, “GANs Trained by a Two Time-Scale Update Rule Converge to a Local Nash Equilibrium,” ArXiv170608500 Cs Stat, Jan. 2018, Accessed: Mar. 13, 2021. [Online]. Available: http://arxiv.org/abs/1706.08500.

[26]	C. Szegedy, V. Vanhoucke, S. Ioffe, J. Shlens, and Z. Wojna, “Rethinking the Inception Architecture for Computer Vision,” ArXiv151200567 Cs, Dec. 2015, Accessed: Mar. 14, 2021. [Online]. Available: http://arxiv.org/abs/1512.00567.

[27]	D. P. Kingma and J. Ba, “Adam: A Method for Stochastic Optimization,” ArXiv14126980 Cs, Dec. 2014, Accessed: Apr. 23, 2019. [Online]. Available: http://arxiv.org/abs/1412.6980.

[28]	X. Dong, Z. Yu, W. Cao, Y. Shi, and Q. Ma, “A survey on ensemble learning,” Front. Comput. Sci., vol. 14, no. 2, pp. 241–258, Apr. 2020, doi: 10.1007/s11704-019-8208-z.

[29]	H. Tang, D. Xu, N. Sebe, and Y. Yan, “Attention-Guided Generative Adversarial Networks for Unsupervised Image-to-Image Translation,” ArXiv190312296 Cs, Aug. 2019, Accessed: Mar. 14, 2021. [Online]. Available: http://arxiv.org/abs/1903.12296.

[30]	Y. Yu, X. Li, and F. Liu, “Attention GANs: Unsupervised Deep Feature Learning for Aerial Scene Classification,” IEEE Trans. Geosci. Remote Sens., vol. 58, no. 1, pp. 519–531, Jan. 2020, doi: 10.1109/TGRS.2019.2937830.

[31]	Z. Yuan et al., “SARA-GAN: Self-Attention and Relative Average Discriminator Based Generative Adversarial Networks for Fast Compressed Sensing MRI Reconstruction,” Front. Neuroinformatics, vol. 14, p. 611666, Nov. 2020, doi: 10.3389/fninf.2020.611666.

[32]	H. Zhang, I. Goodfellow, D. Metaxas, and A. Odena, “Self-Attention Generative Adversarial Networks,” ArXiv180508318 Cs Stat, Jun. 2019, Accessed: Mar. 14, 2021. [Online]. Available: http://arxiv.org/abs/1805.08318.

[33]	Q. Chen and V. Koltun, “Photographic Image Synthesis with Cascaded Refinement Networks,” in 2017 IEEE International Conference on Computer Vision (ICCV), Venice, Oct. 2017, pp. 1520–1529, doi: 10.1109/ICCV.2017.168.

[34]	T. Kaneko, K. Hiramatsu, and K. Kashino, “Generative Attribute Controller with Conditional Filtered Generative Adversarial Networks,” in 2017 IEEE Conference on Computer Vision and Pattern Recognition (CVPR), Honolulu, HI, Jul. 2017, pp. 7006–7015, doi: 10.1109/CVPR.2017.741.

[35]	P.-Y. Laffont, Z. Ren, X. Tao, C. Qian, and J. Hays, “Transient attributes for high-level understanding and editing of outdoor scenes,” ACM Trans. Graph., vol. 33, no. 4, pp. 1–11, Jul. 2014, doi: 10.1145/2601097.2601101.

\section{supplementary  materialst}

\subsection{Re-design dataset}
The data set is collected based on three steps: first, scarping the web for all urban street-level images that are in pair and show a before and after implementation of a given urban policy. Second, sorting these images based on the quality and resolutions of images for training, nevertheless, what has been applied is a viable and realistic urban intervention. This has been asserted based on expert knowledge in the field of urban development, in addition to the realism of the implementation, in case the images introduced are conceptual images, than actual realistic images. For instance, adding trees in a sidewalk (whether this implementation is an implemented case or a design proposal case). Third, images are sorted based on defining a set of categories that could the variety of urban policies implemented in the images, whereas images that do not belong to the common policies or policies classes that have not enough images are excluded. The different policies that are implemented to image A can be  defined as:  

1. Adding cycle lane: It refers to when an applied intervention is limited or include adding a cycle lane.

2. Adding greenery: It refers to when the applied intervention is limited towards adding greenery for an already existing open space, or to the street design 

3. Adding a pedestrian-only zone: It refers to when the applied intervention transforms a road space into a pedestrian zone without access to vehicles. This intervention may include adding greenery.

4. Adding sidewalks: It refers to when the applied intervention transforms a road space include sidewalks by adding them, or by extending them.

5. Façade painting: It refers to painting a wall or building façade without changing its design (no geometric transformation).

6. Façade remodelling: It refers to transforming a building or its façade design to a new design (geometric transformation).

7. Open space remodelling: It refers to re-design an open space or no man’s land to a new design.

8. Road maintenance: It refers to maintaining road surface by either pavement replacement, pothole removal, treefall or garbage removal.

These eight policies represent the translation between image A to image B, however, It worth mentioning that during implementations and training the policy classifier (P), these eight policies represented the images in Group A (the images that need intervention), whereas, the images in group B are used as a null category as no further policy need to be applied as a ninth class of policy categories which can be defined as: 

9. No policy: It refers to the transformed images, where no additional urban intervention is needed.

Besides applying a null category class for the dataset, during implementation and training, a data augmentation strategy has been used. The dataset has been augmented by applying horizontal flipping without any cropping. 



\subsection{Further results and evaluations}

Figure 8 shows our results for both models at resolutions of 1024 X 1024 and 1536 X 1536 respectively. The overall outcomes of both models show a high perceptual realism which in many cases represent similar outcomes, however, there is a couple of perceptual observations that can be made. At 1024 X 1024 resolution, the model tends to lose accuracy at far objects that are not within the ROI of the implemented policies without observation of random noises in the generated image. At 1036 X 1036 resolution, we observe more details learned in smaller regions of the image within and outside the ROI, however, random noises can be found. 

\begin{figure*}
\begin{center}
\includegraphics[width=0.95\linewidth]{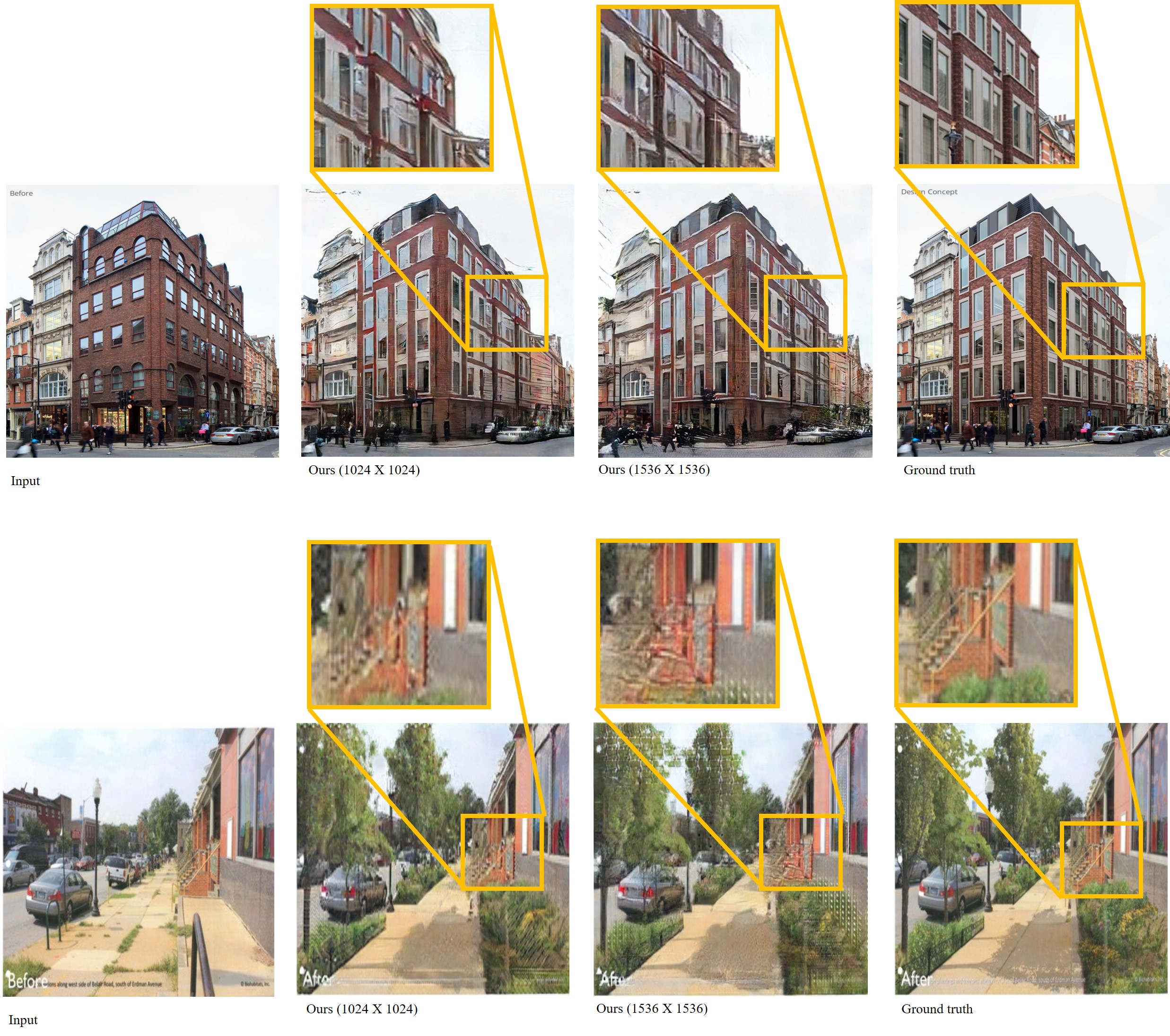}
\end{center}
   \caption{Further results of the proposed model highlighing the variations of the learned tasks}
\label{fig:results1}
\end{figure*}


The following figures (Figure 9-12) shows a wide range of image syntheses generated by our introduced model (1024 X 1024) and the Pix2Pix. By visual inspections, the results of our model surpass the generated images by the Pix2Pix model in all types of intervention policies. The figure shows that the images generated by Pix2Pix are suffering from a limitation of fine details, in addition to over-fitness in the learned features that lead to pattern repetition across different input images regardless of the type of intervention needed.

\begin{figure*}
\begin{center}
\includegraphics[width=1\linewidth]{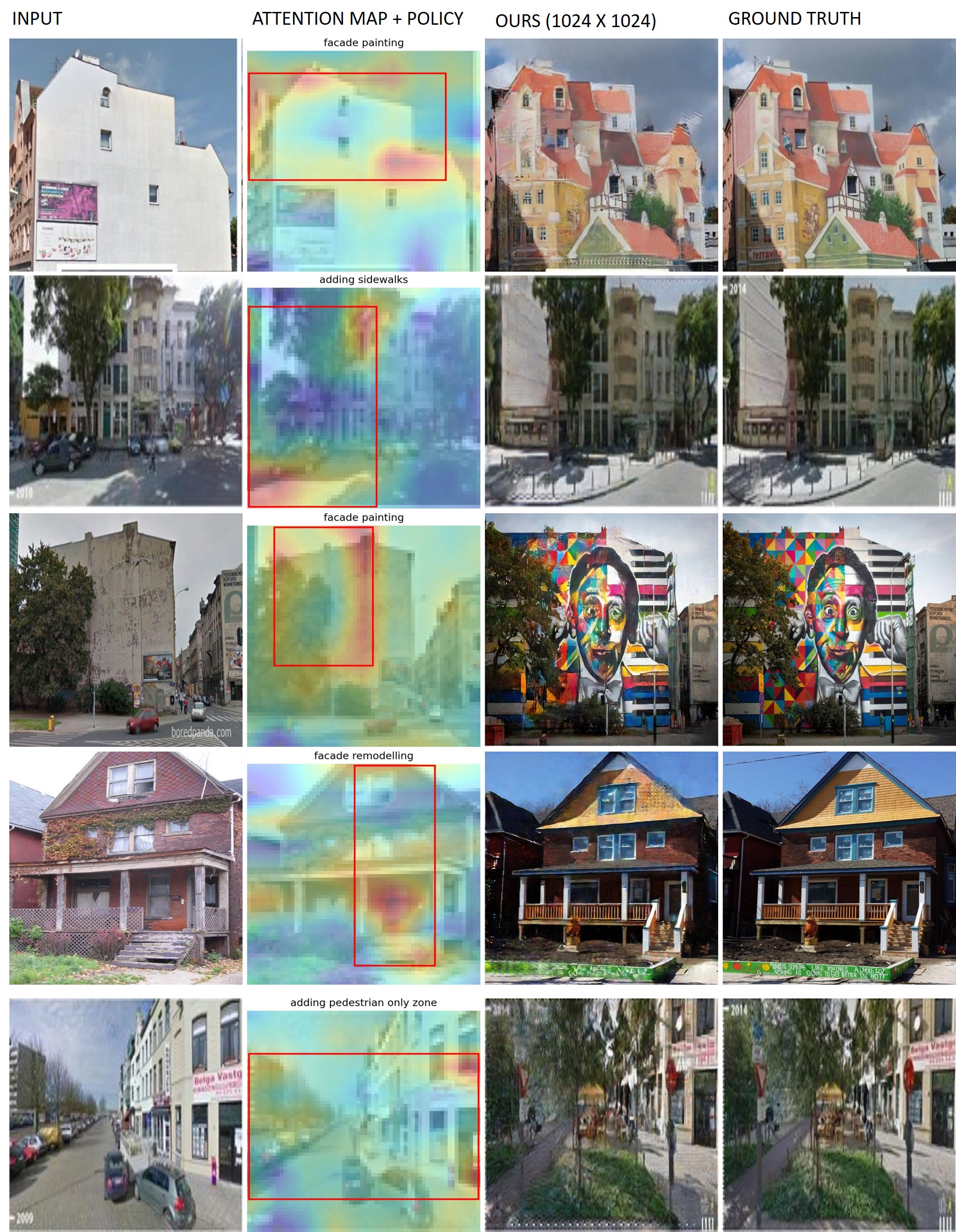}
\end{center}
   \caption{Further results of the proposed model highlighing the variations of the learned tasks}
\label{fig:results1}
\end{figure*}


\begin{figure*}
\begin{center}
\includegraphics[width=1\linewidth]{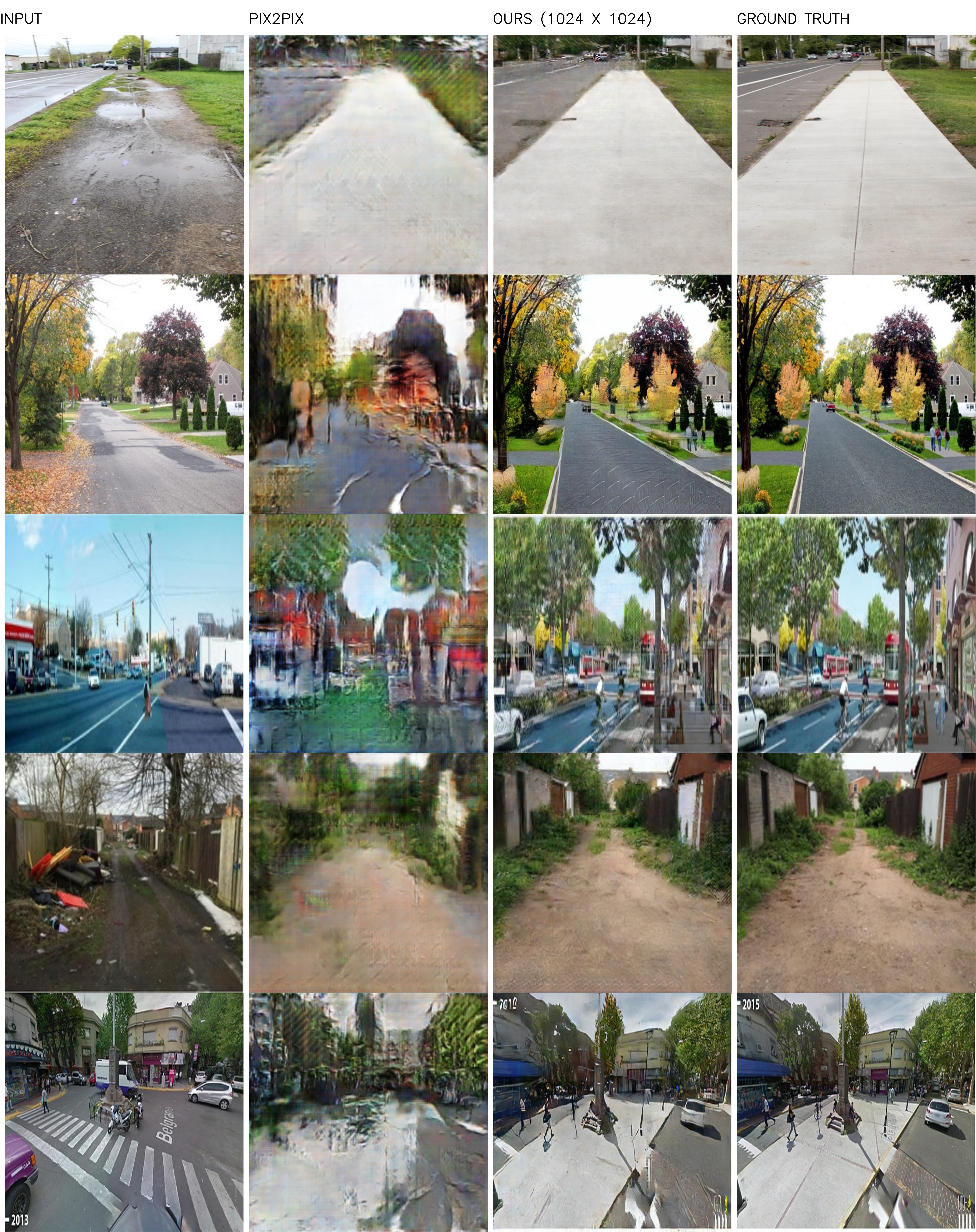}
\end{center}
   \caption{Comparing our results (1024 X 1024) with Pix2Pix model.}
\label{fig:results2}
\end{figure*}


\begin{figure*}
\begin{center}
\includegraphics[width=1\linewidth]{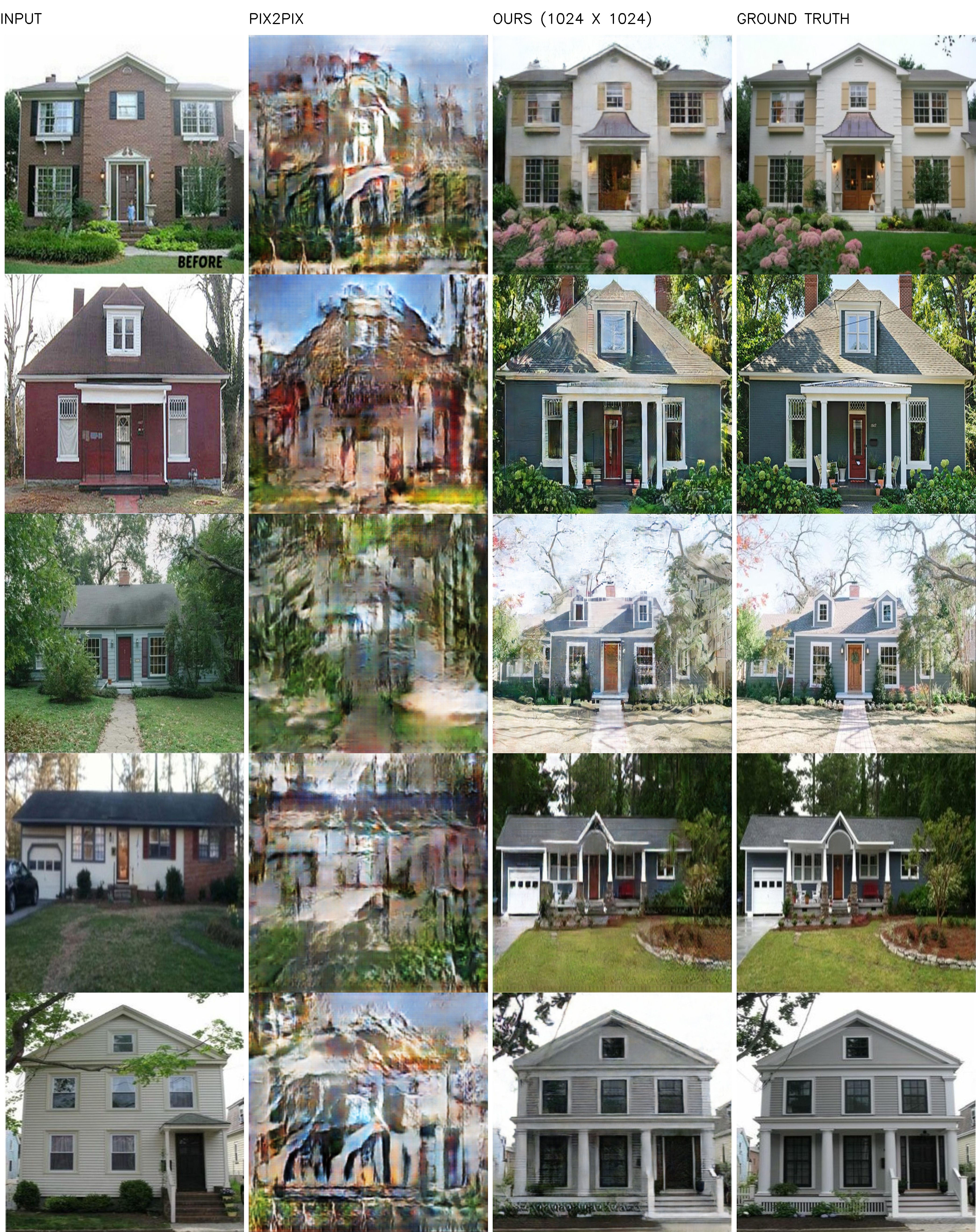}
\end{center}
   \caption{Comparing our results (1024 X 1024) with Pix2Pix model.}
\label{fig:results2}
\end{figure*}


\begin{figure*}
\begin{center}
\includegraphics[width=1\linewidth]{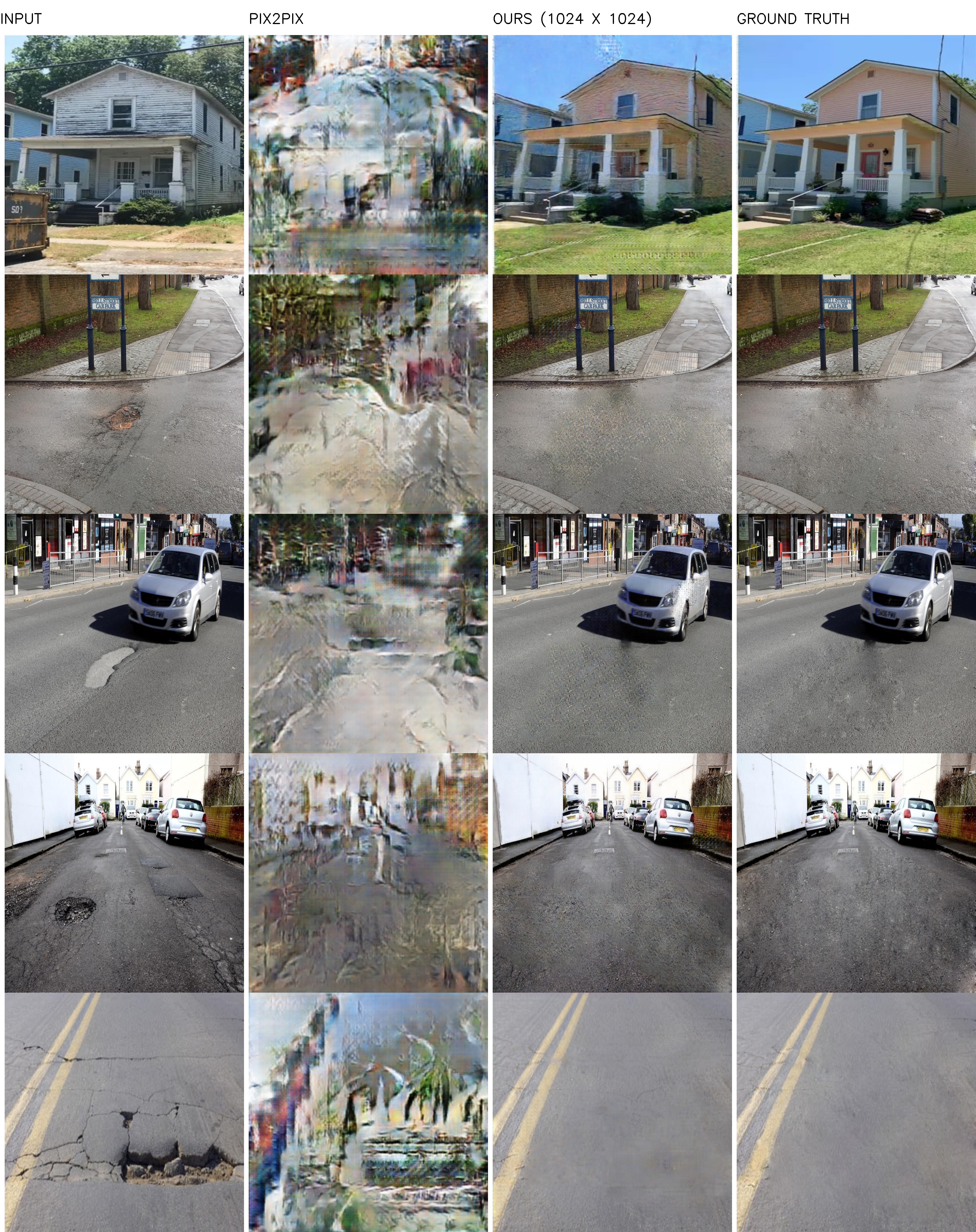}
\end{center}
   \caption{Comparing our results (1024 X 1024) with Pix2Pix model.}
\label{fig:results2}
\end{figure*}


\subsection{Limitations and future work}

There are still several ways for future improvement of the introduced method and its applicability in generating policies for intervention at a city scale. First, regarding methodology, we assume that shifting from ensemble learning towards an adversarial multi-task could lead to a better generalisation. However, the trade-off remains in finding shared learning layers and objective loss between the different introduced tasks, which is an ongoing challenge, in addition to the limitations in computational resources that may increase exponentially when shifting the stated issue to a multitask adversarial problem. On the other hand, decoding the images after gaussian sampling over a latent space could lead to the production of a variety of design interventions for a given input image and a given policy. Last, applying the introduced method to a dataset of a given city (i.e. Google StreetView or data generated by CCTV or cameras deployed in cities) to generate an intervention map and policy suggestions could exemplify the significance of the introduced method.

\end{document}